\documentclass[runningheads]{llncs}

\usepackage{eccv}

\usepackage{eccvabbrv}

\usepackage{graphicx}
\usepackage{booktabs}
\usepackage{xcolor}
\usepackage{multirow}
\usepackage{tikz}
\usepackage{pifont}
\usepackage[skip=2pt]{caption}

\newcommand{\method}[0]{MixCompress}

\usepackage[accsupp]{axessibility}  

\usepackage[pagebackref,breaklinks,colorlinks,citecolor=eccvblue]{hyperref}

\usepackage{orcidlink}

\begin{document}

\title{MixCompress: Mixture of Experts for Variable Rate Learned Image Compression} 

\titlerunning{MixCompress}

\author{Calvin-Khang Ta* \orcidlink{0000-0001-7191-7981} \and
Praneet Singh* \orcidlink{0009-0005-8437-5241} \and
Tong Shao \orcidlink{0009-0003-7485-3663} \and Peng Yin \orcidlink{0000-0001-5423-6359}}

\authorrunning{C-K. Ta et al.}

\institute{Dolby Laboratories, USA \\
\email{\{czta, pusing, tshao, pyin\}@dolby.com}}

\maketitle
\def\thefootnote{*}\footnotetext{Equal Contribution}\def\thefootnote{\arabic{footnote}}
\begin{abstract}
Learned image compression (LIC) is bottlenecked by the need to store independent models for each rate-distortion operating point. Existing variable bit-rate (VBR) methods aim to reduce this overhead via dense parameter modulation, but forcing a shared backbone to approximate divergent mappings causes severe feature entanglement. Specifically, low-rate smoothing gradients inherently conflict with the preservation of high-frequency textural details, leading to sub-optimal performance. To resolve this, we propose \method{}, a unified VBR framework based on sparse structural specialization. While sparsely gated Mixture-of-Experts (MoE) routing successfully mitigates gradient conflict, it operates on a fixed computational budget. To address the increased representational demands of higher bit-rates we introduce a Mixture-of-Depths (MoD) extension to dynamically scale model capacity. Combined with Conditional Auxiliary Transforms (CAT) for dynamic sub-band energy modulation, our hierarchical framework effectively dynamically scales capacity. Extensive evaluations demonstrate that \method{} not only matches individually optimized single-rate baselines but can even surpass them, establishing a new Pareto frontier for computationally efficient image coding.  

\end{abstract}

\section{Introduction}

\begin{figure}[t]
\centering
    \includegraphics[width=0.65\linewidth]{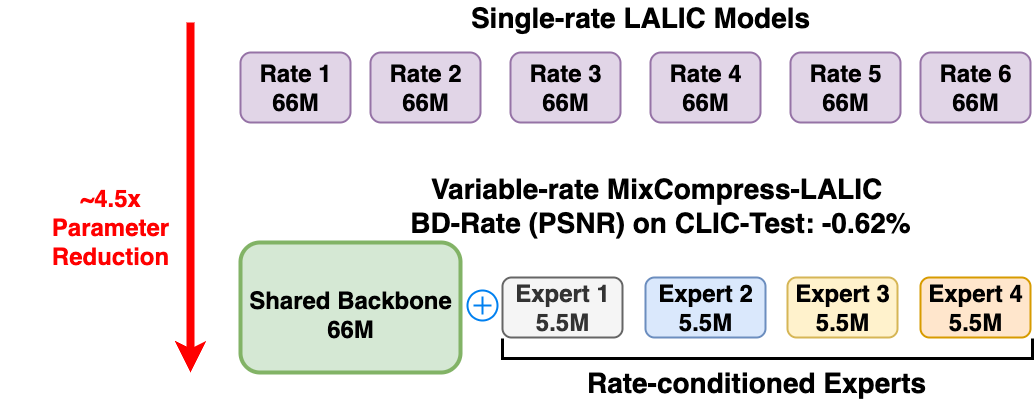}
    \caption{Parameter Comparison. Traditional approaches require separate models for each rate point (top), while \method{} shares 75\% of parameters across all rates with only rate-specific expert modules (bottom). The anchor is the single rate LALIC.}
    \label{fig:param_comparison}
    \vspace{-1pt}
\end{figure}


Learned image compression (LIC) has emerged as a powerful alternative to conventional codecs by jointly optimizing nonlinear transforms and entropy models under a rate–distortion objective \cite{balle2017end}. Hierarchical hyper-priors \cite{balle2018variational}, joint auto-regressive modeling \cite{minnen2018joint}, and refined context modeling \cite{cheng2020learned,he2021checkerboard,he2022elic} progressively closed the gap to traditional codecs. More recent Transformer and linear-attention architectures \cite{liu2023learned,li2024frequencyaware,feng2025linear} further expanded representational power while maintaining practical decoding complexity, establishing LIC as a high-performance learned alternative for image coding. Despite these advances, state-of-the-art LIC models require separate models for each operating point, leading to increased training costs, storage overhead, and deployment complexity. 

Variable-rate LICs aim to address this limitation by supporting multiple operating points within a single unified model. Existing approaches rely on conditioning a shared dense backbone\cite{choi2019conditional,yang2020variable} via latent gain scaling\cite{cui2021agvae} or global affine transformations\cite{tong2023qvrf}. While effective for unified deployment, these methods underperform their single rate optimized counterparts. This performance gap can be attributed to a fundamental optimization flaw: dense parameter modulation forces the single shared backbone to approximate divergent rate-distortion mappings. At the extremes of the rate-distortion trade-off, dense parameter sharing induces severe feature entanglement. Specifically, the parameters responsible for preserving high-frequency textural details at peak bit-rates are subjected to destructive smoothing gradients from low bit-rate optimization. This results in gradient conflict and creates a strict capacity bottleneck.

To overcome this limitation, we introduce \method{}, a unified conditional framework that combines structural specialization with rate-aware representation learning. At the core of our approach is the use of sparsely gated Mixture-of-Experts (MoE) or Mixture-of-Depths (MoD) modules within key bottlenecks of the analysis transform, synthesis transform, and hyper-prior networks. Rather than forcing a shared backbone to compromise across the rate-distortion curve, the routing mechanism provides gradient conflict mitigation. By isolating the gradients from low bit-rate samples we shield the high bit-rate experts from learning suboptimal features. This allows for the model to more effectively model the entire rate-distortion operating curve.

We complement this structural routing with Conditional Auxiliary Transforms (CAT), a rate-aware wavelet shortcut that modulates sub-band energy as an explicit function of the Lagrange multiplier $\lambda$. While MoE and MoD specialize structure, CAT enables feature energy compaction—introducing rate-dependent behavior without altering the entropy model or arithmetic coding pipeline.

By aligning conditional capacity allocation directly with the rate–distortion objective, \method{} effectively decouples the optimization trajectories of competing rate points. Extensive experiments on the Kodak, CLIC, and Tecnick datasets demonstrate our architecture not only closes the variable-rate performance gap, but significantly outperforms state-of-the-art single-rate baselines, including LALIC\cite{feng2025linear}, LIC-TCM\cite{liu2023learned}, and FAT\cite{li2024frequencyaware}. Importantly, our variable-rate models are trained for the same duration as one single-rate model, yet replaces multiple independently trained single-rate checkpoints, substantially reducing both checkpoint storage and total parameter footprint. \method{} thus establishes a new Pareto frontier for practical neural image compression, achieving superior representational density through sparse conditional computation without a proportional increase in inference cost (Fig.~\ref{fig:param_comparison}).

\section{Related Works}

\subsection{Learned Image Compression}

Modern learned image compression (LIC) is based on end-to-end optimized nonlinear transform coding frameworks \cite{balle2017end,balle2018variational}, commonly enhanced with hyper-prior and auto-regressive context models to estimate latent distributions\cite{balle2018variational, minnen2018joint}. Recent work has improved representation capacity through Transformers \cite{liu2023learned,li2024frequencyaware}, linear attention mechanisms \cite{feng2025linear}, and auxiliary wavelet shortcuts (AuxT) that accelerate training via lightweight parallel de-correlation branches \cite{li2025on}. Despite these advances, most LIC methods remain optimized for a single rate point, requiring separate model parameters for different operating bitrates.

\subsection{Variable-Rate Compression}

Variable-rate LIC enables multiple rate-distortion (R-D) operating points within a single model. Conditional autoencoders \cite{choi2019conditional} and modulated autoencoders \cite{yang2020variable} condition the network on the Lagrange multiplier to enable coarse rate control. Gain-based approaches such as the asymmetric gained VAE \cite{cui2021agvae} allow continuous bit-rate adaptation via latent scaling, while quantization-aware frameworks \cite{tong2023qvrf} and scalable decoding methods \cite{ma2022deepfgs} further enhance flexibility.

While these approaches simplify deployment, they largely rely on conditioning or latent modulation within a shared dense backbone. As a result, unified models often underperform individually optimized single-rate counterparts, revealing a gap between parameter sharing and per-rate specialization. Importantly, most methods adapt representations but not model structure, limiting conditional capacity allocation across bit-rate regimes.


\subsection{Mixture-of-Experts and Conditional Computation}

Mixture-of-experts (MoE) architectures enable conditional computation by routing inputs to specialized expert subnetworks \cite{jacobs1991moe}. Sparsely gated MoE layers activate only a subset of experts per sample, dramatically increasing effective capacity without proportional computational cost \cite{shazeer2017moe}. Large-scale systems such as GShard and Switch Transformer \cite{lepikhin2020gshard,fedus2021switch} depict the scalability of conditional routing, and vision MoE models \cite{riquelme2021vmoe} extend these benefits to image tasks. In image compression, mixture-based representations have appeared in alternative coding paradigms \cite{fleig2022smoe}, but sparsely gated MoE has not been integrated into hierarchical transform-based LIC, particularly for variable-rate modeling.

Compared to prior variable-rate methods based solely on conditioning or gain modulation, MoE introduces structural specialization through conditional parameter activation. When applied to the analysis transform, synthesis transform, and hyperprior networks, MoE enables bit-rate-dependent capacity allocation throughout the compression pipeline. Extending this idea with depth-wise adaptivity (MoD) further enhances conditional computation. Together with recent advances in structured energy compaction such as AuxT, these developments motivate a unified variable-rate image compression framework.

\section{Methodology}

\begin{figure}[t!]
    \centering
    \includegraphics[width=0.80\linewidth]{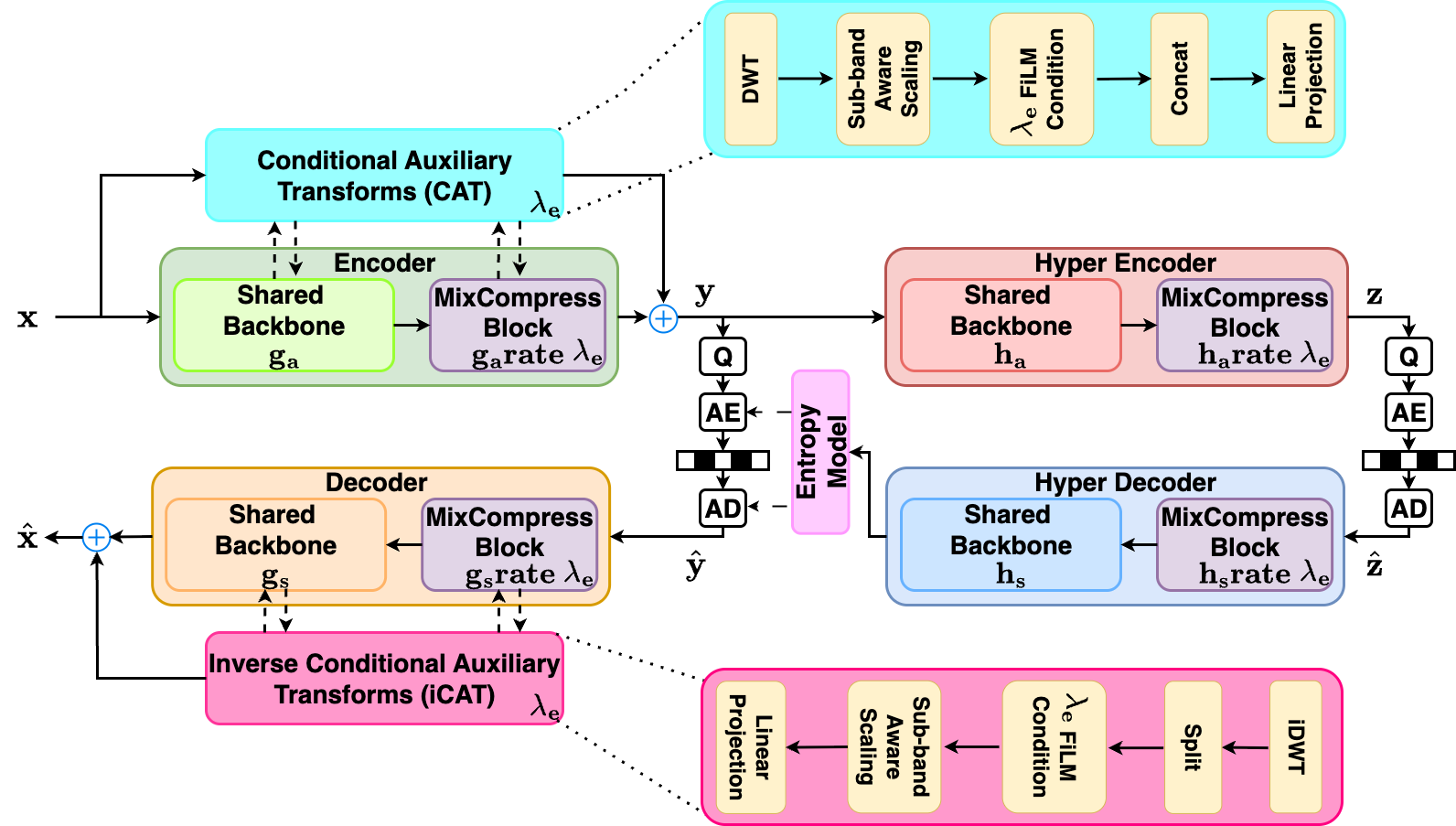}
\caption{Overview of the \method{} architecture. We adapt a LIC codec and replace dense bottleneck blocks in the analysis, synthesis, and hyper-prior transforms with rate-conditioned ($\lambda_e$) \method{} blocks. Sub-band energy compaction is further modulated through Conditional Auxiliary Transforms (CAT/iCAT), enabling dynamic adaptation across rate–distortion operating points.}
    \label{fig:main_block}
\end{figure}
\subsection{Preliminaries: Variable Rate Learned Image Compression}
\noindent\textbf{Variable Learned Image Compression Formulation:}
Modern learned image compression relies on end-to-end optimized non-linear transform coding framework\cite{balle2017end, balle2018variational}. An input image $x$ is mapped to a latent representation $y$ via an analysis transform $g_a$, and is reconstructed as $\hat{x}$ via a synthesis transform $g_s$. The latent $y$ is further mapped to a hyper-latent $z$ through the hyper-encoder $h_a$. Both the latent and hyper-latent are quantized and denoted as $\hat{y}$ and $\hat{z}$ respectively. In order to entropy code $\hat{z}$ we use the standard factorized prior\cite{balle2018variational}. The hyper-decoder($h_s$) then estimates the entropy parameters $\mu,\sigma$ from $\hat{z}$ which is refined with the space-channel context entropy model\cite{he2022elic}. A standard single rate model optimizes a parameter set $\theta$ for the rate-distortion objective:
\begin{align}
    \min_\theta \mathcal{L}_\lambda(\theta) = \mathbb{E}_{x\sim p_{data}}[\lambda D(x,\hat{x}; \theta) + R(\hat{y};\theta) + R(\hat{z};\theta)]
\end{align}
where $D$ is our distortion metric and $\lambda$ is a fixed Lagrange multiplier. In order to support variable-rate control without having a set of parameters for each $\lambda$, unified frameworks sample $\lambda\sim p(\lambda)$ during training, effectively optimizing the expectation over a distribution of tasks: 
\begin{align}
    \min_\theta \mathcal{L}_{VBR}(\theta) = \mathbb{E}_{\lambda \sim p(\lambda),x\sim p_{data}}[\lambda D(x,\hat{x}; \theta) + R(\hat{y};\theta) + R(\hat{z};\theta)]
\end{align}

\noindent\textbf{Limitations of Dense Parameter Modulation:}
To adapt a shared backbone to varying $\lambda$, existing methods rely on dense parameter modulation, such as conditional convolutions\cite{choi2019conditional} or latent scaling\cite{tong2023qvrf}. Given an embedded rate parameter $\lambda_e$, a shared feature map $f$ is modulated via a learned affine transformation $f' = \sigma(MLP_s(\lambda_e)) * \text{Conv2D}(f) + MLP_b(\lambda_e)$
where $\sigma$ is an activation function (e.g., SoftPlus).

While dense conditioning enables variable rate functionality, it consistently under-performs models optimized for single rate points. We hypothesize that this performance gap is due to a multi-objective optimization failure: Gradient Interference. Let $g_\text{low} = \nabla_\theta\mathcal{L}_{\lambda_{low}}$ and $g_\text{high} = \nabla_\theta\mathcal{L}_{\lambda_{high}}$ represent the gradients for updating a shared set of parameters for both low and high bit-rate targets. High bit-rate optimization requires the model preserve high-frequency structural details, whereas low bit-rate optimization strictly penalizes detail in favor of blurrier reconstructions thus lowering the bit-rate. This results in a scenario where the cosine similarity between task gradients is potentially negative\cite{yu2020gradient}.
When cosine similarity is negative the gradients for either task will degrade the model performance for the opposite task. Applying affine transformations in the form of conditional convolutions or latent scaling is not sufficient for resolving this conflict as this only scales the feature magnitude and will force the model to learn a suboptimal compromise between the two rate points. 

\subsection{Conditional Auxiliary Transforms (CAT)}

Improving feature energy compaction has been shown to accelerate training and enhance performance in LIC. Auxiliary linear shortcuts, such as AuxT \cite{li2025on}, leverage wavelet-based decompositions to offload coarse feature de-correlation and sub-band energy modulation from the main nonlinear transform, stabilizing optimization and improving convergence. However, existing shortcuts use static sub-band scaling, limiting flexibility in variable-rate LIC where the optimal energy distribution depends on the target rate: low-rate regimes benefit from stronger compaction, while high-rate regimes require finer detail preservation.

To address this, we introduce \emph{Conditional Auxiliary Transforms (CAT)}, a rate-aware extension that conditions sub-band energy modulation on the Lagrangian multiplier $\lambda$. CAT retains the structural benefits of wavelet shortcuts while enabling adaptive energy reweighting across operating points.

\noindent\textbf{Wavelet Decomposition}:
Given an intermediate feature map
$
\mathbf{P} \in \mathbb{R}^{B \times C \times H \times W},
$
CAT applies a 2D Haar discrete wavelet transform (DWT) as follows:
\begin{equation}
\mathbf{Z} = \mathrm{DWT}(\mathbf{P})
\in
\mathbb{R}^{B \times 4C \times \frac{H}{2} \times \frac{W}{2}}.
\end{equation}
The transform decomposes $\mathbf{P}$ into four frequency subbands
\(\{\mathbf{P}_{LL}, \mathbf{P}_{LH}, \mathbf{P}_{HL}, \mathbf{P}_{HH}\}\),
each in
\(
\mathbb{R}^{B \times C \times \frac{H}{2} \times \frac{W}{2}}
\),
corresponding to low-frequency (LL) and high-frequency
(LH, HL, HH) components \cite{li2025on}. This is then tokenized into
$
\mathbf{U}
=
\mathrm{reshape}(\mathbf{Z})
\in
\mathbb{R}^{B \times \frac{HW}{4} \times 4C}.
$

\noindent\textbf{Rate-Conditioned Subband Aware Scaling:}
Each CAT block maintains learnable scaling vectors for each sub-band
$
\mathbf{s}_{\text{base}}
\in
\mathbb{R}^{4C}.
$ Let
\(
\lambda_e \in \mathbb{R}^{d_\lambda}
\)
be the learned embedding of the selected rate point.
A FiLM head produces rate-conditioned scaling parameters:
\begin{equation}
[\boldsymbol{\gamma}_\lambda,\boldsymbol{\beta}_\lambda]
=
f_{\mathrm{film}}(\lambda_e),
\qquad
\boldsymbol{\gamma}_\lambda,\boldsymbol{\beta}_\lambda
\in
\mathbb{R}^{4C}.
\end{equation}
These rate-conditioned scalars are then used to update the static subband aware scaling parameters of AuxT \cite{li2025on} as follows:
$ \mathbf{s}_\lambda^{\text{enc}} = \boldsymbol{\gamma}_\lambda \odot \mathbf{s}_{\text{base}} + \boldsymbol{\beta}_\lambda$.
Encoder-side modulation is applied token-wise:
$
\hat{\mathbf{U}}
=
\mathbf{U}
\odot
\exp(\mathbf{s}_\lambda^{\text{enc}}),
$
where $\odot$ denotes channel-wise multiplication. A learned linear projection then maps the rate-conditoned wavelet tokens to the shortcut output channels:
$\mathbf{V} = \hat{\mathbf{U}} \mathbf{W}_{\text{enc}}, \qquad \mathbf{W}_{\text{enc}} \in \mathbb{R}^{4C \times C'}$.
Reshaping yields
$
\mathbf{V}
\in
\mathbb{R}^{B \times C' \times \frac{H}{2} \times \frac{W}{2}},
$
which is injected residually at the various locations of the encoder as done in AuxT.

\noindent\textbf{Decoder Symmetry:}
iCAT  mirrors the encoder operations on the decoder side. Given decoder features
\(
\mathbf{Y}
\in
\mathbb{R}^{B \times C' \times H \times W},
\)
we compute
\begin{equation}
\mathbf{U}'
=
\mathrm{reshape}
(\mathbf{Y}\mathbf{W}_{\text{dec}}),
\qquad
\mathbf{W}_{\text{dec}}
\in
\mathbb{R}^{C' \times 4C}.
\end{equation}
A decoder FiLM head produces
\(
\mathbf{s}^{\text{dec}}_\lambda
\in
\mathbb{R}^{4C}
\),
and reciprocal scaling is applied:
$
\tilde{\mathbf{U}}'
=
\mathbf{U}'
\odot
\exp(-\mathbf{s}^{\text{dec}}_\lambda).
$
After reshaping to
\(
\mathbb{R}^{B \times 4C \times \frac{H}{2} \times \frac{W}{2}},
\)
inverse DWT reconstructs the spatial shortcut: $\hat{\mathbf{P}}=\mathrm{IDWT}(\tilde{\mathbf{U}}')$. CAT/iCAT preserves the plug-and-play nature of AuxT, leaving the main transform and entropy model unchanged, while elevating auxiliary branches be fully rate-aware. By conditioning subband energy modulation on $\lambda_e$, our models can dynamically adjust energy compaction across all various rate points.

\subsection{Mixture of Experts}
\begin{figure}[t!]
    \centering
    \includegraphics[width=0.7\linewidth]{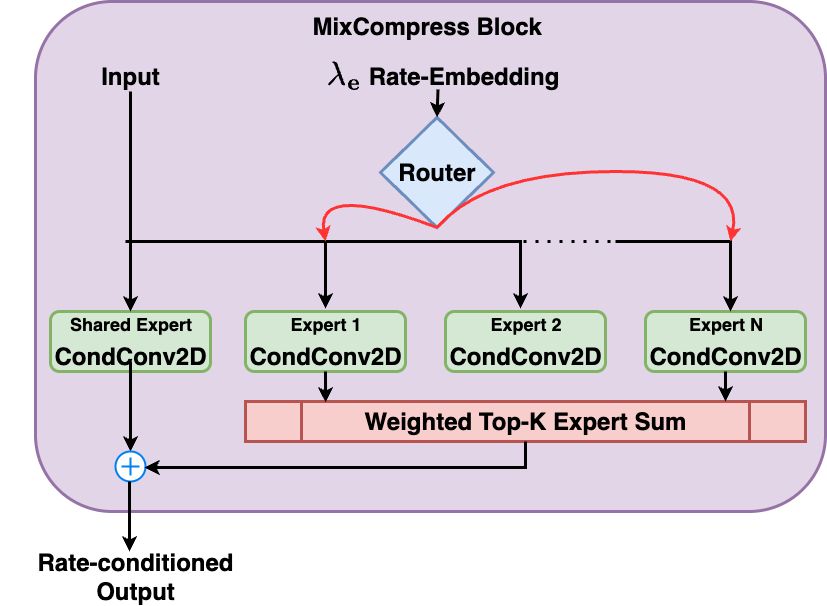}
    \caption{The \method{} block. The input feature is always processed by a shared expert to ensure stable information flow, and is summed with the outputs of the sparsely routed top-K experts, thus mitigating conflicting gradient updates.}
    \label{fig:moe_router}
\end{figure}

To resolve the gradient interference inherent to dense parameter modulation, we introduce \method{}. By replacing the last layer at key bottlenecks with sparsely gated Mixture-of-Experts (MoE) modules, we decouple the optimization trajectories of competing rate points.

\noindent\textbf{Sparse Structural Routing:} Given an input feature $f$ and a target rate embedding $\lambda_e$, a routing network conditioned on $\lambda_e$ generates a probability distribution over $N$ parallel expert networks $\{E_1, E_2, ..., E_N\}$. To ensure that universally useful features get through for all rate-points we maintain a separate shared expert $E_{global}$. The output of the \method{} block is computed as the sum of the shared expert and the top-$K$ selected experts :$f' = E_{global}(f) + \sum_{i\in\mathcal{T}_k}G(\lambda_e)E_i(f)$ where $\mathcal{T}_k$ is the set of top-K expert indices selected by the softmax gating mechanism $G(\lambda_e)$. In our implementation we set $K=2$. Since our router is only conditioned on the $\lambda_e$ we face the high likelihood of expert collapse\cite{chi2022representation,riquelme2021vmoe}. This results in a small subset of experts getting updated as other experts are completely ignored. To encourage balanced expert utilization we employ noise annealing and inject additive Gaussian noise $\epsilon$ into the router logits during training: $G(\lambda_e) = Softmax(TopK(MLP(\lambda_e)+\epsilon))$.
This encourages the model to explore all experts and is gradually reduced to zero over time (schedule details in Sec. \ref{sec:exp}).

\noindent\textbf{Task Conditioned Experts:} A key architectural constraint in our framework is that both the gating networks $G(\lambda_e)$ and the CAT modules depend strictly on the rate embedding $\lambda_e$ which is entirely independent from the input image $x$. Because these modulations are independent of the input data, they do not actually need to be computed at inference. For practical deployment, the exact expert assignment and modulation parameters can be precomputed and cached in a lookup table, introducing zero routing overhead at inference.

\begin{figure}[t!]
    \centering
    \includegraphics[width=.7\linewidth]{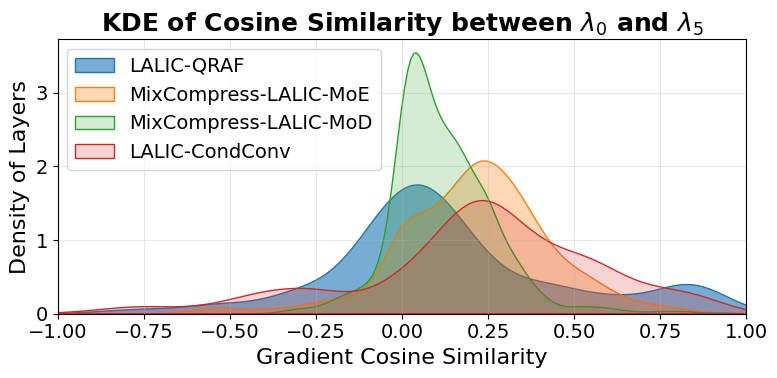}
    \caption{Kernel Density Estimation of layer-wise gradient cosine similarity between our lowest and highest operating points $\lambda_{0}$ and $\lambda_{5}$ on 100 training images. The dense approaches (QRAF\cite{tong2023qvrf}, CondConv\cite{choi2019conditional}) exhibit long negative tails, indicating severe gradient conflict. \method{} is able to mitigate the interference shifting the optimization dynamics towards orthogonality.}
    \label{fig:grad_conflict}
\end{figure}

\noindent\textbf{Gradient Conflict Mitigation:} 
The primary advantage of this sparse structural routing is the capacity to mitigate gradient interference, a well documented optimization failure in multi-task and multi-objective learning\cite{yu2020gradient,sener2018multi}.
Let the parameters of a \method{} block be partitioned in to the shared expert $\theta_{shared}$ and the routed experts $\Theta_{routed} = [\theta_1, ..,\theta_N]$. For a given $\lambda$, the gradient of the loss $\mathcal{L}_\lambda$ decomposes into updates for these respective subspaces. The inner product between the gradients for two widely separated operating points $\lambda_{low}$ and $\lambda_{high}$ is the sum of their subspace inner products: 

\begin{equation}
\begin{aligned}
    \langle g(\lambda_{low}), g(\lambda_{high})\rangle = \langle g_{shared}(\lambda_{low}),g_{shared}(\lambda_{high})\rangle \\
    + \langle g_{routed}(\lambda_{low}),g_{routed}(\lambda_{high})\rangle
\end{aligned}
\end{equation}
The shared expert processes gradients from all operating points and is still subject to gradient conflicts. As shown in prior multi-objective optimization literature\cite{liu2021conflict,sener2018multi}, forcing conflicting updates through shared parameters limits the model's ability to reach the true Pareto frontier. However, the inclusion of a routing mechanism isolates gradient conflicts within conditional capacity based on expert overlap. Let $\mathcal{T}_{low}$  and $\mathcal{T}_{high}$ be the selected experts for their respective rate point. The inner product of the routed subspace is strictly bounded by the intersection of these active experts:
\begin{align}
    \langle g_{routed}(\lambda_{low}), g_{routed}(\lambda_{high})\rangle = \sum_{i\in \mathcal{T}_{low} \cap\mathcal{T}_{high}} \langle \nabla_\theta\mathcal{L}_{low}, \nabla_\theta\mathcal{L}_{high}\rangle
\end{align}
The shared expert $\theta_{shared}$ is still subject to potential gradient conflict but by reducing the volume of parameters subject to conflicting updates, we expect the overall interference to diminish. We empirically validate this in Fig. \ref{fig:grad_conflict}, where \method{} shifts the gradient cosine similarity distribution towards zero, confirming that sparse structural routing mitigates gradient conflicts. Decomposing the conflict by network layers reveals that the encoder is disproportionately affected under dense modulation: $78\%$ and $79\%$ of the encoder layers in QRAF and CondConv respectively exhibit negative  gradient cosine similarity. \method{}-LALIC-MoE and \method{}-LALIC-MoD  reduces this to $31\%$ and $22\%$ respectively. This is consistent with visual quality improvements observed in high-bit rate reconstructions (Additional per layer statistics in Supplemental Material).

\noindent\textbf{Mixture of Depths(MoD) Extension:} While standard MoE frameworks allocate distinct parameters, maximizing fidelity at peak bit-rates often requires additional representational capacity. We extend the proposed framework by progressively increasing expert complexity. In the MoD variant, experts will remain compatible by outputting the same tensor shapes but contain varying depths. We formulate the $i$-th expert as a composite function of $i$ nested transformation blocks (in our case Bi-RWKV blocks) $E_i(f) = (h_i\circ h_{i-1}...h_1)(f)$. By routing high-rate samples to higher complexity experts the model can dynamically expand its representational capacity when required.

\section{Experiments}
\label{sec:exp}

\subsection{Experimental Setup}

We follow the training protocol of LALIC~\cite{feng2025linear} and train all models on the first 400K images of OpenImages~\cite{openimages}. OpenImages provides diverse high-resolution natural content, making it well suited for learned image compression. During training, images are randomly cropped to $256 \times 256$ patches. All models are optimized using Adam with a batch size of 8. Rate--distortion optimization is performed under the MSE objective using the following multipliers: $\lambda \in \{0.0018, 0.0035, 0.0067, 0.0130, 0.0250, 0.0483\}$.

In contrast to single-rate baselines where each $\lambda$ requires an independently trained model, in our approach we jointly learn all operating points within a single unified model by uniformly sampling one $\lambda$ per mini-batch during training. This strategy ensures balanced supervision across rates while eliminating the need for multiple checkpoints. As a result, our approach substantially reduces training cost, storage overhead, and deployment complexity.

Importantly, although our model supports multiple rate points, we follow the same training schedule as LALIC and train for the same total duration as one single-rate model. Training is conducted in three stages. We first train for 40 epochs with a learning rate of $1\times10^{-4}$, followed by 4 epochs at $1\times10^{-5}$. Finally, we fine-tune the model for 4 additional epochs using $512 \times 512$ random crops to improve high-resolution reconstruction quality. All experiments are performed on a single NVIDIA H100 GPU.
\subsection{Model Configuration}

We integrate the proposed framework into two strong single-rate backbones, LALIC~\cite{feng2025linear} and LIC-TCM~\cite{liu2023learned}. 
To ensure a strictly controlled comparison, we preserve the original architecture of each backbone and introduce our variable-rate modules only at the end of the analysis transform, synthesis transform, and hyperprior network, as illustrated in Fig.~\ref{fig:main_block}. 
All remaining components are kept identical to their original implementations.

Variable-rate adaptation is enabled through a 32-dimensional rate embedding $\lambda_e$, which conditions routing decisions within each \method{} block. Each block contains $N$ routed experts with top-$K$ sparse selection, together with one additional always-active shared expert that ensures stable information flow. For the main experiments, we use $N=4$ and $K=2$ for all \method{} variants unless specified otherwise. All experts share identical input–output tensor interfaces, enabling seamless integration without modifying the surrounding architecture. The effect of different expert counts $N$ and routing sparsity levels $K$ is analyzed in the supplementary material.

\noindent\textbf{\method{} Variants:} We evaluate two expert configurations here:
\begin{itemize}
    \item \textbf{MoE (Homogeneous Experts).}
    In both \method{}-LALIC-MoE and \method{}-TCM-MoE, all routed experts share the same internal architecture and differ only in parameters. 
    This setting isolates the effect of parameter specialization under a fixed computational graph.

    \item \textbf{MoD (Progressive-Depth Experts).}
    In \method{}-LALIC-MoD, routed experts progressively stack additional $2\times$ Bi-RWKV\cite{feng2025linear} blocks, forming a shallow-to-deep hierarchy. 
    In \method{}-TCM-MoD, depth is increased more conservatively using $1\times$ convolutional \cite{liu2023learned} blocks to control overall model size.
\end{itemize}

\noindent\textbf{CAT Shortcuts:} We enable CAT and iCAT in the encoder and decoder, respectively. Four conditional wavelet shortcuts are inserted symmetrically in both branches of the \method{} architecture, providing rate-adaptive auxiliary pathways without altering the primary codec structure.

\noindent\textbf{Routing Stabilization:} To encourage balanced expert utilization in our rate-conditioned routers during training, we inject additive Gaussian noise into the router logits using a linearly annealed schedule, $\sigma(e)=\sigma_0 \cdot \max\left(1-\frac{e}{E_{\text{anneal}}},0\right)$, where $e$ denotes the current epoch. We use $\sigma_0=1.0$ and $E_{\text{anneal}}=20$ by default. The noise is therefore large during the early epochs to promote exploration, gradually decays over training, and becomes zero after epoch 20, allowing stable expert specialization. To validate the effectiveness of this stabilization strategy, we provide comprehensive expert utilization and routing statistics in the Supplemental Material, confirming that noise annealing mitigates catastrophic routing collapse and maintains diverse expert utilization across most of the network.

\subsection{Evaluation Datasets and Metrics}

We evaluate on four widely used image compression benchmarks: Kodak~\cite{kodak1993lossless} (24 images at $768 \times 512$ resolution), Tecnick~\cite{tecnick} (100 images at $1200 \times 1200$ resolution), and the CLIC Professional validation and test sets~\cite{clic2021workshop}, containing 41 and 60 high-resolution images (up to 2K resolution), respectively. Rate-distortion performance is measured using bits per pixel (BPP) and peak signal-to-noise ratio (PSNR). We report BD-rate savings computed using PSNR as the distortion metric and VTM-23.1 \cite{VTM} as the anchor. All R--D curves are generated using the same set of $\lambda$ values across all methods to ensure strict fairness.

\subsection{Rate-Distortion Performance}
\begin{figure}[ht]
    \centering
    \includegraphics[width=.7\linewidth]{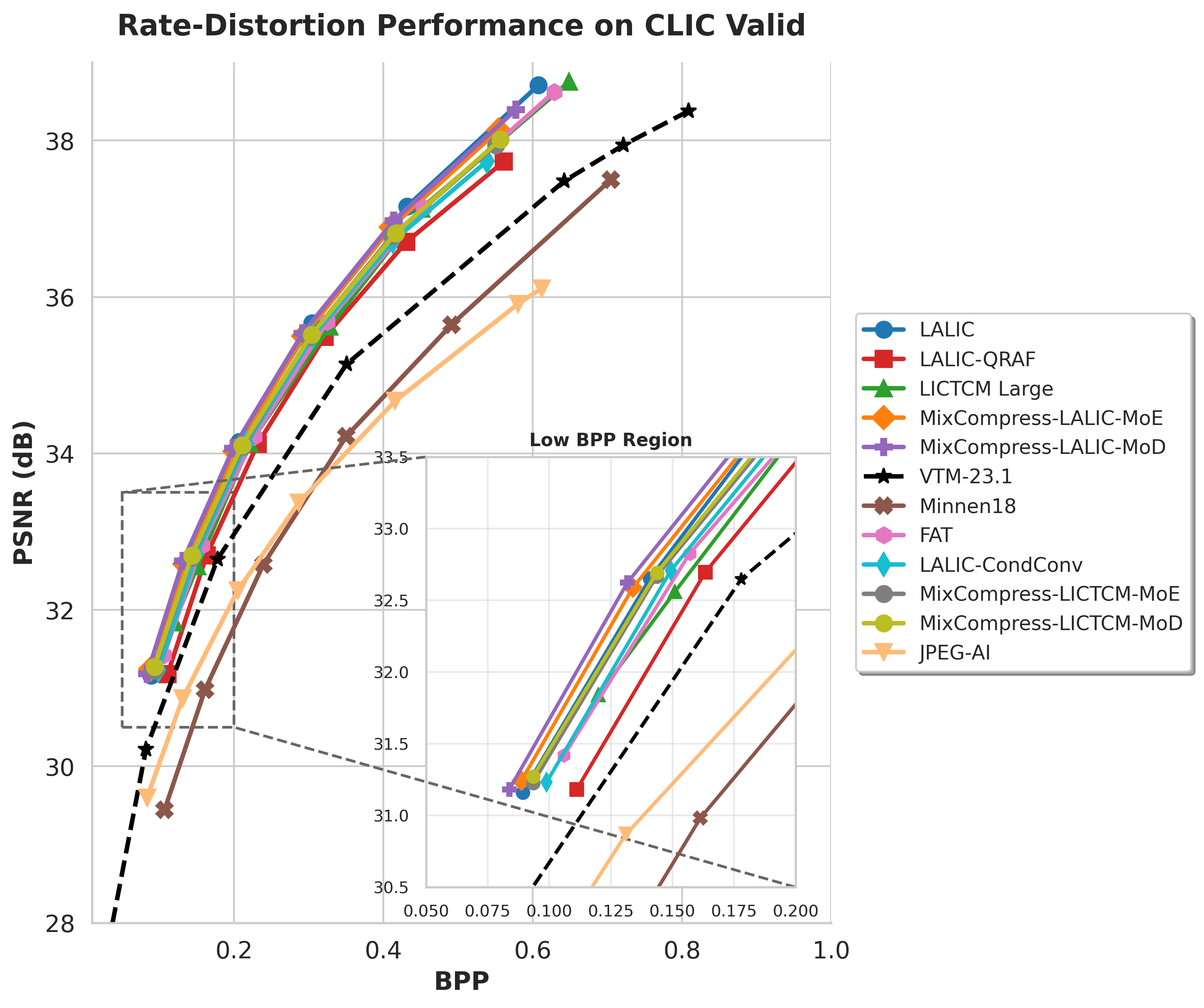}
    \caption{RD curves demonstrating the improvement achieved by \method{} on CLIC Validation. JPEG-AI\cite{jpegai} is included for completeness, though note that direct comparison is not straightforward as it is optimized for perceptual metrics rather than distortion metrics. Additional RD curves at full resolution for CLIC Validation, CLIC Test, Kodak, and Tecnick, evaluated on both PSNR and MS-SSIM, are provided in the Supplemental Material.}
    \label{fig:rd_curves}
\end{figure}

\newcommand{\cmark}{\ding{51}}%
\newcommand{\xmark}{\ding{55}}%

\begin{table}[]
\centering
\caption{Rate–distortion comparison in terms of BD-Rate(PSNR)\%. Results are computed using VTM-23.1 as the anchor. The \textcolor{green}{best}, \textcolor{red}{second-best}, and \textcolor{blue}{third-best} results for each dataset are highlighted. Variable bit-rate (VBR) methods indicated by \cmark}
\begin{tabular}{lccccc}
\hline
\multirow{2}{*}{Method} & \multirow{2}{*}{VBR} & \multicolumn{4}{c}{BD-Rate (PSNR)} \\ \cline{3-6}
 & & Kodak & CLIC-V & CLIC-T & Tecnick \\ \hline

VTM-23.1 \cite{VTM} & \cmark & 0.00 & 0.00 & 0.00 & 0.00 \\

Hyper-prior \cite{balle2017end} & \xmark & 27.09 & 34.27 & 35.38 & 35.88\\
Minnen18 \cite{minnen2018joint} & \xmark & 6.89 & 2.95 & 0.98 & 3.30 \\
Cheng20-Parallel \cite{cheng2020learned} & \xmark & -1.77 & -4.119 & -5.45 & -3.52 \\
ELIC \cite{he2022elic} & \xmark & -8.95 & -12.22 & -14.35 & -15.07 \\

TCM \cite{liu2023learned} & \xmark & -16.06 & -18.63 & -19.70 & -20.41 \\

FAT \cite{li2024frequencyaware} & \xmark & \textcolor{blue}{-17.12} & -19.07 & -20.04 & -23.07 \\
LALIC \cite{feng2025linear} & \xmark & \textcolor{red}{-18.81} & \textcolor{blue}{-23.96} & \textcolor{blue}{-26.43} & \textcolor{blue}{-26.03} \\

\hline
LALIC-QRAF & \cmark & -9.32 & -13.22 & -15.24 & -14.36\\

LALIC-CondConv & \cmark & -15.12 & -19.22 & -22.05 & -21.60 \\

\method{}-LALIC-MoE 
& \cmark
& \textcolor{red}{-18.81} 
& \textcolor{red}{-24.16} 
& \textcolor{red}{-26.80} 
& \textcolor{red}{-27.05} \\

\method{}-LALIC-MoD 
& \cmark
& \textcolor{green}{-20.12} 
& \textcolor{green}{-25.73} 
& \textcolor{green}{-28.71} 
& \textcolor{green}{-28.88} \\

\hline
\end{tabular}
\label{tab:bd-rate}
\end{table}

Table~\ref{tab:bd-rate} reports BD-rate results relative to VTM-23.1 \cite{VTM}. We evaluate MixCompress against two categories of methods: (i) single-rate baselines, where each operating point requires an independently trained model, and (ii) variable-rate (VBR) baselines, which support multiple rate points within a single model.

\noindent\textbf{Comparison with Single-Rate Models:} \method{} variants consistently match or surpass their corresponding single-rate checkpoint across all benchmarks despite being trained as a single unified model for the same total duration, where each rate point effectively sees only 1/6th of the training iterations compared to a dedicated single-rate checkpoint. On CLIC Valid, CLIC Test, and Tecnick datasets, \method{}-LALIC-MoD achieves $-25.73\%$, $-28.71\%$, and $-28.88\%$ BD-rate (PSNR) reductions, exceeding the individually optimized baseline by a significant margin. Even the homogeneous \method{}-LALIC-MoE variant improves upon single-rate LALIC on CLIC and Tecnick. On the lower-resolution Kodak dataset, \method{}-LALIC-MoD surpasses the single-rate LALIC anchor while \method{}-LALIC-MoE closely matches it. Notably, \method{}-LALIC-MoD outperforms all other single-rate architectures in our comparison, including FAT\cite{li2024frequencyaware} and LIC-TCM\cite{liu2023learned}, across every benchmark while only requiring a single trained checkpoint rather than six.

\noindent\textbf{Comparison with Variable-Rate Models:} We re-implement two representative dense-modulation VBR strategies, QRAF\cite{tong2023qvrf} and Conditional Convolutions\cite{choi2019conditional} on the LALIC backbone using identical training data, schedule, and evaluation protocol. Both dense VBR baselines incur substantial performance degradation. In contrast, \method{}-LALIC-MoE effectively closes this variable-rate gap, and \method{}-LALIC-MoD surpasses the single-rate counterpart entirely. The pattern is consistent across all four benchmarks with dense modulation methods sacrifice between $3.7$-$12.0$ percentage points of BD-rate relative to the single-rate LALIC. We attribute \method{}'s performance to the gradient conflict mitigation enabled by sparse structural routing, validated empirically in Table. \ref{tab:bd-rate}. We note that earlier gain-based VBR methods\cite{cui2021agvae} have been shown to underperform QRAF\cite{tong2023qvrf} under comparable settings; we therefore treat QRAF as the stronger dense-modulation reference and omit\cite{cui2021agvae} from our comparison for brevity. 

\begin{figure}[ht!]
    \centering
    \includegraphics[width=\linewidth]{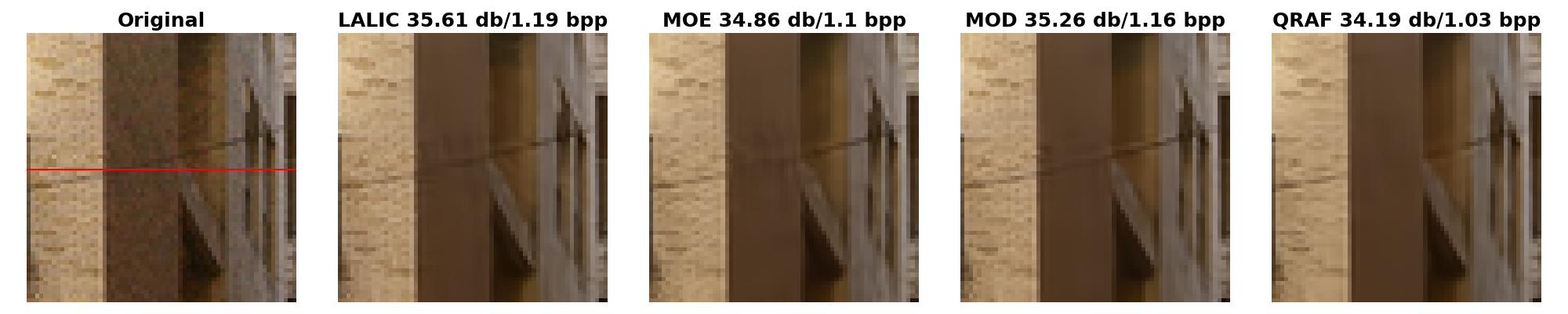}\\
    \caption{Visual Comparison best viewed zoomed in. Here we cropped out a section of an image from the CLIC validation split. LALIC-QRAF at the highest quality is unable to reconstruct the cable (denoted by the red line in the original) present in the original image while the single-rate baseline LALIC as well as our multi-rate versions are capable at the same rate-point.}
    \label{fig:visual_comparison}
\end{figure}
\vspace{-1em}
\noindent\textbf{Visual Comparison:} In Fig. \ref{fig:visual_comparison} a localized impact of gradient conflict on high-frequency structural elements can be seen. At comparable high bit-rates, the dense QRAF\cite{tong2023qvrf} fails to reconstruct fine details, such as the thin cable crossing the building. This over-smoothing effect is a result of competing objectives where low bit-rate gradients favor this behavior while high bit-rate gradients would want to preserve the cable. By structurally isolating these gradients both \method{}-MoE and \method{}-MoD are able to successfully preserve details.

\noindent\textbf{MoE vs. MoD:} MoD variants consistently outperform MoE, suggesting that depth-adaptive experts provide advantages beyond parameter-only specialization by adapting representational capacity to bit-rate demand. Cross-backbone generalization to LIC-TCM\cite{liu2023learned}, visual comparisons, and continuous rate interpolation at unseen $\lambda$ values are validated in the Supplemental Material.

\vspace{1em}
\begin{figure}[t!]
    \centering
    \includegraphics[width=\linewidth]{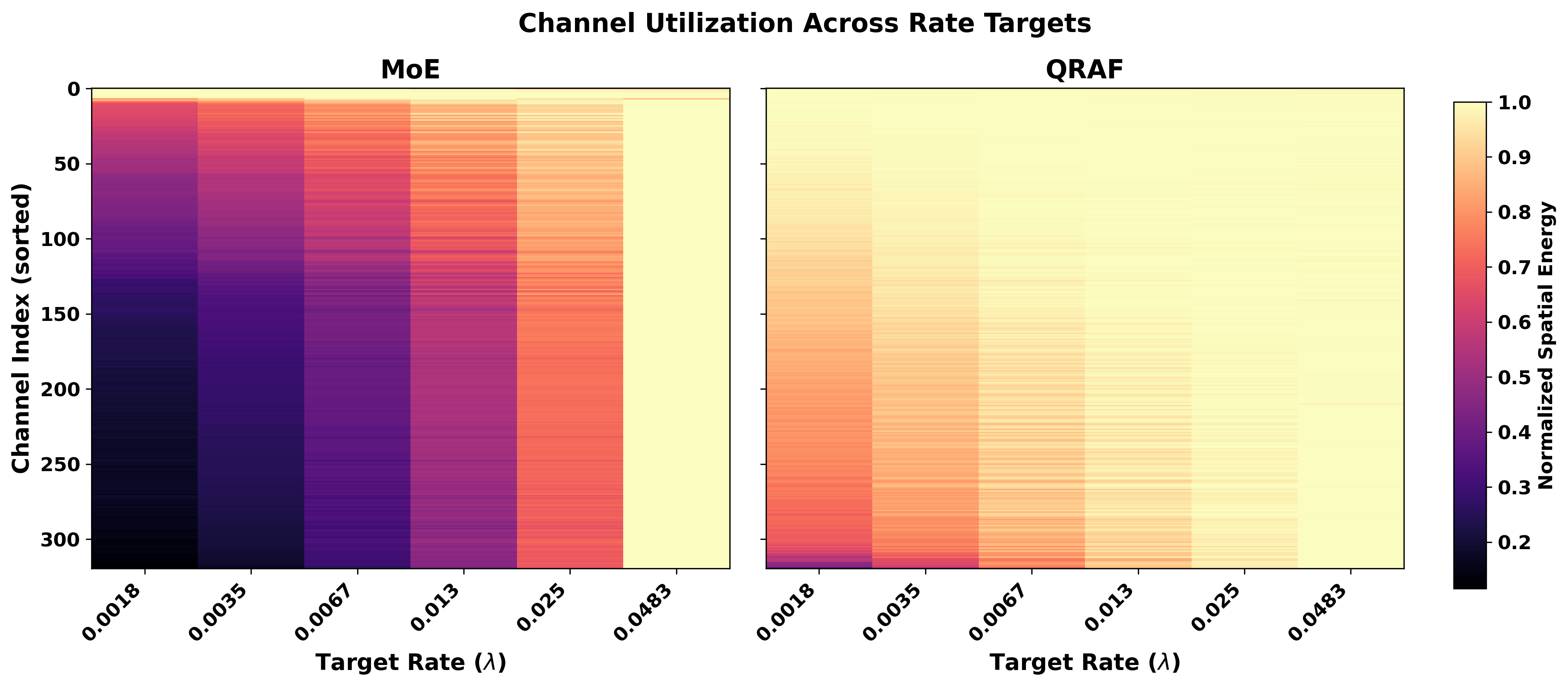}
    \caption{Channel utilization heatmaps across different rates $\lambda$. The dense baseline QRAF\cite{tong2023qvrf}(right) forces highly overlapping, entangled parameter usage across all rate points. In contrast, \method{}(left) dynamically reallocates capacity, sharing channels between adjacent rates for smoother transitions between rate-points while decoupling the parameters for low and high bit-rates to minimize gradient conflict.}
    \label{fig:allocation}
\end{figure}
\vspace{-2em}
\subsection{Latent Analysis}
\noindent\textbf{Gradient Conflict Mitigation:} To empirically validate our hypothesis that dense conditioning induces optimization interference, we analyze the layer-wise gradient cosine similarity between two extreme rate points ($\lambda_{low}$ and $\lambda_{high}$) in Fig. \ref{fig:grad_conflict}. The dense baselines QRAF\cite{tong2023qvrf} and Cond-Conv\cite{choi2019conditional} both exhibit long negative tails in their distributions, indicating severe gradient conflict where updates for low-rate abstraction will directly degrade parameters for high fidelity reconstruction. By structurally isolating  these trajectories, \method{} successfully shifts the dynamics towards orthogonality, heavily increasing the gradient similarity around zero and mitigating destructive interference.

\noindent\textbf{Capacity Allocation:} The gradient conflict mitigation is physically realized in the channel utilization strategy. As visualized in Fig. \ref{fig:allocation}, a method like QRAF forces a dense, entangled parameter utilization across all target rates. In contrast, \method{} is able to more gradually transition between rate points and more effectively decouples active parameters from low and high rate points. At the high bit-rates, QRAF exhibits signs of behavior of low bit-rate models which can be observed in the overly smoothed images. The severe under-performance of LALIC-QRAF is attributable to two compounding failure modes specific to this backbone. QRAF adapts bit-rate by scaling the latent exclusively at the quantization boundary, meaning the sole rate-differentiating signal is quantization noise variance. However, Bi-RWKV's LayerNorm\cite{ba2016layer} normalizes per-token variance at each block, effectively squashing this signal before it can influence downstream representations. As a result, both the entropy model and synthesis transform become rate-blind, learning a single compromise across all operating points. This incompatibility is further compounded by the gradient interference inherent to dense modulation, which remains unresolved. \method{} avoids both failure modes by routing latents through structurally distinct pathways, producing representations that differ in direction rather than magnitude. This is further corroborated by the RD-curves (Fig. \ref{fig:rd_curves}) where we can see QRAF has a severely limited range in both PSNR and BPP. We provide qualitative visual reconstructions demonstrating \method{}'s superior preservation of high-frequency at higher bit-rates in the Supplemental Material. 

\subsection{Computational and Deployment Efficiency}

\begin{table}[t!]
    \centering
        \caption{We report various metrics measured during inference. Note that for MoE/MoD models we report both the active number of parameters/total number of parameters. The memory usage is including all experts loaded into memory.}
\begin{tabular}{lccccc}
\hline
\multirow{1}{*}{Method}                   & \multirow{1}{*}{Enc(s)} & \multirow{1}{*}{Dec(s)} & \multirow{1}{*}{Mem(G)} & \multirow{1}{*}{FLOPs(G)} & \multirow{1}{*}{Params(M)} \\ 

\hline
LALIC\cite{feng2025linear}                & 0.234                   & 0.184                   & 0.875                   & 283.39                   & 66.13         \\
\hline
\method-MoE                               & 0.252                   & 0.187                   & 1.037                   & 311.99                   & 77.86/88.56           \\ 
\method-MoD                     & 0.257                   & 0.190                   & 1.262                   & 392.02                   & 120.22/156.55         \\ 
\hline
\end{tabular}
    \label{tab:complexity}
\end{table}

We provide the computational requirements and inference latency of our framework in Table \ref{tab:complexity}. Because our routers and CAT modulations are only conditioned on the rate embedding $\lambda_e$, the expert choices and modulation parameters only need to be computed once. As a result, those network parameters do not need to be kept in memory during inference. We report the optimized cached results and we can see there is only a slight increase in Encoding and Decoding time. This significantly smaller than maintaining an ensemble of a single rate model such as LALIC\cite{feng2025linear} would require around 397 million parameters in storage. In contrast, \method-MoD our largest model, covers the same rate-distortion curve with 156 million parameters with only 120 million active per pass. 

\section{Ablation Studies}

We conduct controlled ablations on the LALIC backbone to systematically evaluate each design component of \method{}. 
All variants are trained under identical settings and evaluated using the R-D protocol in Sec.~\ref{sec:exp}. 
Table~\ref{tab:ablation_detailed} reports BD-rate differences relative to the VTM-23.1 \cite{VTM} anchor.

We begin with a unified variable-rate model implemented via conditional convolution (Cond-Conv), where convolutional weights are modulated by the rate embedding $\lambda_e$. 
Although this dense modulation strategy leads to clear performance improvement over VTM-23.1 on all benchmark datasets, this simple weight modulation is insufficient to preserve the performance of independently trained single-rate models from Table. \ref{tab:bd-rate}.

Introducing AuxT shortcuts partially mitigates this gap, reducing the BD-rate loss relative to the single-rate baseline and indicating that frequency-domain shortcuts help stabilize parameter sharing across rates. Replacing AuxT with our rate-aware CAT transforms further improves performance, showing that explicit rate-conditioned subband modulation is more effective than static auxiliary transforms for variable-rate modeling.

Next, we replace the  Cond-Conv layers with our MoE blocks. This narrows the performance gap to the single-rate models on Kodak and surpasses them on CLIC and Tecnick  (Table. \ref{tab:bd-rate}). This depicts that sparse expert specialization is more effective than dense weight modulation for variable-rate LIC. Finally, utilizing our MoD blocks consistently yields the strongest BD-Rate performance across all datasets. By enabling routing to control both parameter selection and representational capacity, \method{}-MoD effectively adapts model capacity across rates and outperforms the independently trained single-rate baselines.

\begin{table}[t]
    \centering
    \caption{Component-wise ablation of MixCompress. All models trained with variable rate control with 4 experts. Additional encoder only results in the Supplemental.}
    \label{tab:ablation_detailed}
    \resizebox{.65\linewidth}{!}{
    \begin{tabular}{lcccc}
        \toprule
        \textbf{Configuration} & \textbf{Kodak} & \textbf{CLIC-V} & \textbf{CLIC-T} & \textbf{Tecnick} \\
        \midrule
              VTM-23.1 \cite{VTM}                         & 0 & 0 & 0 & 0 \\
        Variable-rate (Cond-Conv)     & -15.15 & -19.27 & -22.09 &  -21.53\\
            \quad + AuxT \cite{li2025on}            & -17.06      & -20.76 & -23.37 & -23.50 \\
        \quad + CAT             &   -17.89  & -21.92 & -24.23 & -24.41 \\
        \quad + MoE (N=4, K=2)  & -18.81 & -24.16 & -26.80 &  -27.05 \\
        \quad + MoD (N=4, K=2) & \textbf{-20.12} & \textbf{-25.73} &  \textbf{-28.71} & \textbf{-28.88} \\
        \bottomrule
    \end{tabular}
    }
\end{table}

\section{Conclusion}

We propose \method{}, a unified variable-rate learned image compression framework that overcomes the gradient interference inherent to dense parameter modulation. By integrating sparsely gated Mixture-of-Experts (MoE) and Mixture-of-Depth (MoD) modules, alongside Conditional Auxiliary Transforms (CAT), our architecture dynamically partitions network capacity based on the target rate $\lambda$. This structural specialization successfully isolates competing rate-distortion objectives allowing for more efficient variable-rate models. Extensive evaluations on Kodak, CLIC, and Tecnick datasets demonstrate that \method{} consistently matches and often surpasses individually trained single-rate baselines. By demonstrating that conditional structural allocation is fundamentally more effective than dense weight modulation, \method{} establishes a new Pareto frontier for practical, multi-rate neural image compression.

%
%
\clearpage
\bibliographystyle{splncs04}
\bibliography{main}
\end{document}